\documentclass[sigconf]{acmart}

\newif\ifanon
\makeatletter
\if@ACM@anonymous
\anontrue
\fi
\makeatother

\usepackage[shortlabels,inline]{enumitem}
\setlist{nolistsep} 
\setlist[itemize]{leftmargin=*} 

\usepackage{todonotes}

\usepackage{mathtools}
\DeclarePairedDelimiter\floor{\lfloor}{\rfloor}

\graphicspath{{figures/}}

\def \questionname {Q}
\newcommand{\qref}[1]{\questionname\ref{#1}}
\newcounter{qcounter}
\newenvironment{question}[1]{\refstepcounter{qcounter}\textbf{(\questionname\theqcounter)} #1}{}
\usepackage{subfigure}

\newcommand{\eg}{e.g., }

\newcommand{\ie}{i.e., }

\newcommand{\figref}[1]{Fig.~\ref{#1}}    
\newcommand{\Figref}[1]{Figure~\ref{#1}}  

\newcommand{\secref}[1]{\S\ref{#1}}
\newcommand{\equref}[1]{Eq.~(\ref{#1})}

\setcopyright{acmcopyright}
\copyrightyear{2019}
\acmYear{2019}
\acmDOI{10.1145/3360322.3360992}

\acmConference[BuildSys 2019]{BuildSys 2019: ACM International Conference on Systems for Energy-Efficient Buildings, Cities, and Transportation}{November 13--14, 2019}{New York, NY, USA}
\acmBooktitle{BuildSys 2019: ACM International Conference on Systems for Energy-Efficient Buildings, Cities, and Transportation,
 November 13--14, 2019, New York, NY, USA}
\acmISBN{978-1-4503-7005-9/19/11}

\def\tslot{\Delta t^\text{slot}}
\def\tdepart{\Delta t^\text{depart}}
\def\tcharge{\Delta t^\text{charge}}
\def\tflex{\Delta t^\text{flex}}

\def\Tflex{T_\textit{flex}}
\def\Eflex{E_\textit{flex}}

\def\Smax{S_\text{max}}
\def\Hmax{H_\text{max}}
\def\Nmax{N_\text{max}}
\def\Cbau{C_\text{BAU}}
\def\Copt{C_\text{opt}}
\def\CRLold{C_\text{RL,old}}
\def\CRLnew{C_\text{RL,updated}}
\def\Cheur{C_\text{heur}}

\def\xtot{\textbf{x}^\textrm{total}_{s}}
\def\action{\textbf{u}_s}

\def \ea {\textit{et al.\ }}

\begin{document}

\title[Optimized cost function for DR coordination of multiple EV charging stations using reinforcement learning]{Optimized cost function for demand response coordination of multiple EV charging stations using reinforcement learning}

\author{Manu Lahariya}
\affiliation{%
  \institution{IDLab, Ghent University -- imec}}
\email{manu.lahariya@ugent.be}

\author{Nasrin Sadeghianpourhamami }
\affiliation{%
  \institution{IDLab, Ghent University -- imec}}
\email{nasrin.sadeghianpourhamami@ugent.be}

\author{Chris Develder}
\affiliation{%
  \institution{IDLab, Ghent University -- imec}}
\email{chris.develder@ugent.be}


\begin{abstract}%
  Electric vehicle (EV) charging stations represent a substantial load with significant flexibility. Exploitation of that flexibility in demand response (DR) algorithms becomes increasingly important to manage and balance demand and supply in power grids. Model-free DR based on reinforcement learning (RL) is an attractive approach to balance such EV charging load. We build on previous research on RL, based on a Markov decision process (MDP) to simultaneously coordinate multiple charging stations. However, we note that the computationally expensive cost function adopted in previous research leads to large training times, which limits the feasibility and practicality of the approach. 
  We therefore propose an improved cost function which essentially forces the learned control policy to always fulfill any charging demand that does not offer any flexibility. We rigorously compare the newly proposed batch RL fitted Q-iteration implementation with the original (costly) one, using real world data. Specifically, for the case of load flattening, we compare the two approaches in terms of 
  \begin{enumerate*}[(i)]
  \item the processing time to learn the RL-based charging policy, as well as
  \item overall performance of the policy decisions in terms of meeting the target load for unseen test data.
  \end{enumerate*}
  The performance is analyzed for different training periods and varying training sample sizes.
  In addition to both RL policies' performance results, we provide performance bounds in terms of both
  \begin{enumerate*}[(i)]
  \item an optimal all-knowing strategy, and
  \item a simple heuristic spreading individual EV charging uniformly over time.
  \end{enumerate*}

\end{abstract}



\maketitle

\section{Introduction}
\label{sec:Intro}

    
    
    Demand response (DR) algorithms are pivotal to ensure demand-supply balance in smart grids with intermittent renewable energy resources and new loads (\eg electric vehicles, EVs).
    In traditional approaches for coordinating EV charging~\cite{hu2016electric}, DR is cast as an optimization problem (\eg model predictive control, MPC). However, this approach requires accurate models (\eg of user behavior, energy demand, flexibility that is available to exploit) which have uncertainty associated with them. Furthermore, such approaches do not generalize from one scenario to the other.
    
    Aforementioned challenges are tackled with recent data-driven DR algorithms, where the charging coordination problem is cast as a time-series decision making problem and is formulated using Markov decision process (MDP) with unknown system dynamics. Reinforcement learning (RL) is then used to estimate the optimum charging coordination policy (\eg \cite{claessens2013RLEV}). 
    For a recent overview of RL in DR, we refer to \cite{VazquezCanteli2019}.
    In terms of objectives, different DR targets have been addressed, including
    \begin{enumerate*}[(i)]
    \item reducing electricity costs,
    \item maximizing profits for the provider, and
    \item load balancing in the grid.
    \end{enumerate*}
    For example, Chis \ea\cite{chics2016reinforcement} propose a reduction in long term cost for user for charging a single EV. 
    
    Previous research \cite{nasrin2018mdp} formulated an MDP for a set of EV charging stations, aiming at a model-free DR approach for EV charging stations to exploit time flexibility provided by users. In~\cite{nasrin2019journal}, a refined MDP and experimental performance evaluation is provided, thus giving a proof-of-principle of adopting RL for jointly coordinating charging for a set of EVs. This approach can simultaneously control multiple EV charging at once, in contrast to \cite{chics2016reinforcement}, which optimizes charging for just a single EV. 
    The MDP definition scales independently of the number of charging stations ($\Smax$) and number of maximum cars ($\Nmax$), and thus can be easily deployed in multiple scenarios. 
    While the learned RL policy's performance demonstrated effectiveness in terms of meeting the DR objective, it comes at the cost of requiring a large set of experiences (past data), long training periods and computational power.
    
    This paper extends the previous work in \cite{nasrin2019journal} by improving the RL implementation: 
    we reduce the computational complexity and dataset requirements in the MDP definition and RL training through
    \begin{enumerate*}[(i)]
        \item an updated cost function, and
        \item a reduced state-action space in MDP, resulting in a smaller exploration dataset.
    \end{enumerate*}
    The rest of the paper presents the following contributions:
    \begin{itemize}[topsep=0pt]
    \item We propose an updated MDP with a new cost function (\secref{subsec:mdp});
    \item We train RL policies for both the original \cite{nasrin2019journal} and updated cost functions (using the algorithm summarized in \secref{subsec:batchRL});
    \item Simulation experiments (\secref{sec:experimentsetup}) to evaluate the policies and answer the following questions (\secref{sec:results}):
    \begin{itemize}[label=,topsep=0pt]
		\item \begin{question}\label{q:traintime:reduction} What reduction of training time and computational complexity does the new cost function achieve?\end{question}
		\item \begin{question}\label{q:triantime:variance} How does varying the parameters of input training data impact the training time?\end{question}
		\item \begin{question}\label{q:performance} Does the updated cost function affect the resulting performance achieved by the RL policy?\end{question}
		\item 
		\begin{question}\label{q:etflex} How much of the offered flexibility does the RL policy use?\end{question}
	\end{itemize}
    \end{itemize}


\section{Algorithm}
\label{sec:algo}
A Markov decision process (MDP) is defined by 
\begin{enumerate*}[(i)]
  \item a finite state space,
  \item an action space,
  \item a cost (or rewards) function for taking a particular action, given a state.
\end{enumerate*}
The next subsections summarize
\begin{enumerate*}[(i)]
\item the MDP for jointly coordinate charging a set of EVs, and 
\item a batch reinforcement learning algorithm for training the policy.
\end{enumerate*}

\subsection{Markov Decision Process (MDP)}
\label{subsec:mdp}
    
    \subsubsection*{State:}
    To create a state for the set of connected EVs, at a given timeslot $t$,
    we assume to know for each EV 
    \begin{enumerate*}[(i)]
    \item the time left until it departs ($\tdepart$), and
    \item its charging requirement, which we quantify as the number of timeslots it needs to charge ($\tcharge$), thus assuming the same constant charging rate for each EV.
    \end{enumerate*}
    Hence, the state representation is of the form $s=(t,\textbf{x}_s)$, where $t$ is the timeslot (\ie $t \in \{1,\dots,\Smax\}$) and  $\textbf{x}_s$ is the aggregate demand of all EVs in the system. Formally, $\textbf{x}_s$ is an $\Smax{} \times \Smax{}$ matrix, where $\Smax{}$ is the maximum timeslots in the horizon. The element of $\textbf{x}_s$ at position $(i, j)$ counts the fraction of EV charging stations that have a connected car in the corresponding $(\tdepart, \tcharge)$ bin, \ie for which $i = \tdepart$ and $j = \tcharge$.

    \subsubsection*{Action:}
    Per set of EVs that have a particular flexibility, the action dictates whether or not to charge them.
    We note that EVs with the same flexibility are positioned along the diagonals of the state matrix $\textbf{x}_s$: the flexibility, defined as the amount of time shifting we can apply in the charging process, is indeed given by $\tflex = \tdepart - \tcharge$.
    We indicate the number of EVs on each diagonal of $\textbf{x}_s$ as $\xtot(d)$ with $d = 0,\dots,\Smax-1$, where $\xtot(0)$ counts the EVs on the main diagonal, $\xtot(d)$ on the upper $d^\textrm{th}$ diagonal, and $\xtot(-d)$ on the lower $d^\textrm{th}$ diagonal of $\textbf{x}_s$.
    The action $\textbf{u}_s$ vector thus defines an action per diagonal, and for each of them states
    what fraction of the diagonal's EVs to charge.
 
    \subsubsection*{Cost Function:} 
    We aim to flatten the aggregate EV charging load, while ensuring that every EV is fully charged before departing.
    The original proposition in \cite{nasrin2019journal} combined both aims as separate parts in the cost function for transitioning from $s$ to $s'$ by taking action $\action$:
    \begin{equation}
        C(s,\action,s')_\textrm{old} \triangleq C^{\text{demand}}(\textbf{x}_s,\action) + C^{\text{penalty}}(\textbf{x}_{s'}),
    \label{eq:RL:OldCost} 
    \end{equation}
    where $ C^{\text{demand}}(\textbf{x}_s,\action)$ is the (quadratic) power consumption from all connected EVs in the decision timeslot. 
    $C^{\text{penalty}}(\textbf{x}_{s'})$ is the penalty for unfinished charging, defined to be higher than simultaneously charging all EVs. 
    Thus, $C^{\text{penalty}}$ 
    is activated when a car would move to below the main diagonal (where $\tcharge{} > \tdepart{}$), to ensure fully charging all EVs, including those without flexibility.
    
    Our newly defined cost for taking action  $\action$ to get from cost state $s$ to $s'$ amounts to the charging power demand cost only: 
    \begin{equation}
        C(s,\action,s')_\textrm{updated} \triangleq C^{\text{demand}}(\textbf{x}_s,\action),
    \label{eq:RL:UpdatedCost} 
    \end{equation}
    In the updated cost definition, we no longer define the penalty term:
    we impose the policy to a priori charge all cars without any flexibility (\ie those on the main diagonal, where indeed $\tcharge = \tdepart$, thus $\tflex = 0$), rather than having it learn to do that from experiencing a high penalty cost. 
    This results in a reduced state-action space and significantly faster training for the model.
    
    \begin{figure}[!t]
    	\centering
    	\includegraphics[width=\columnwidth]{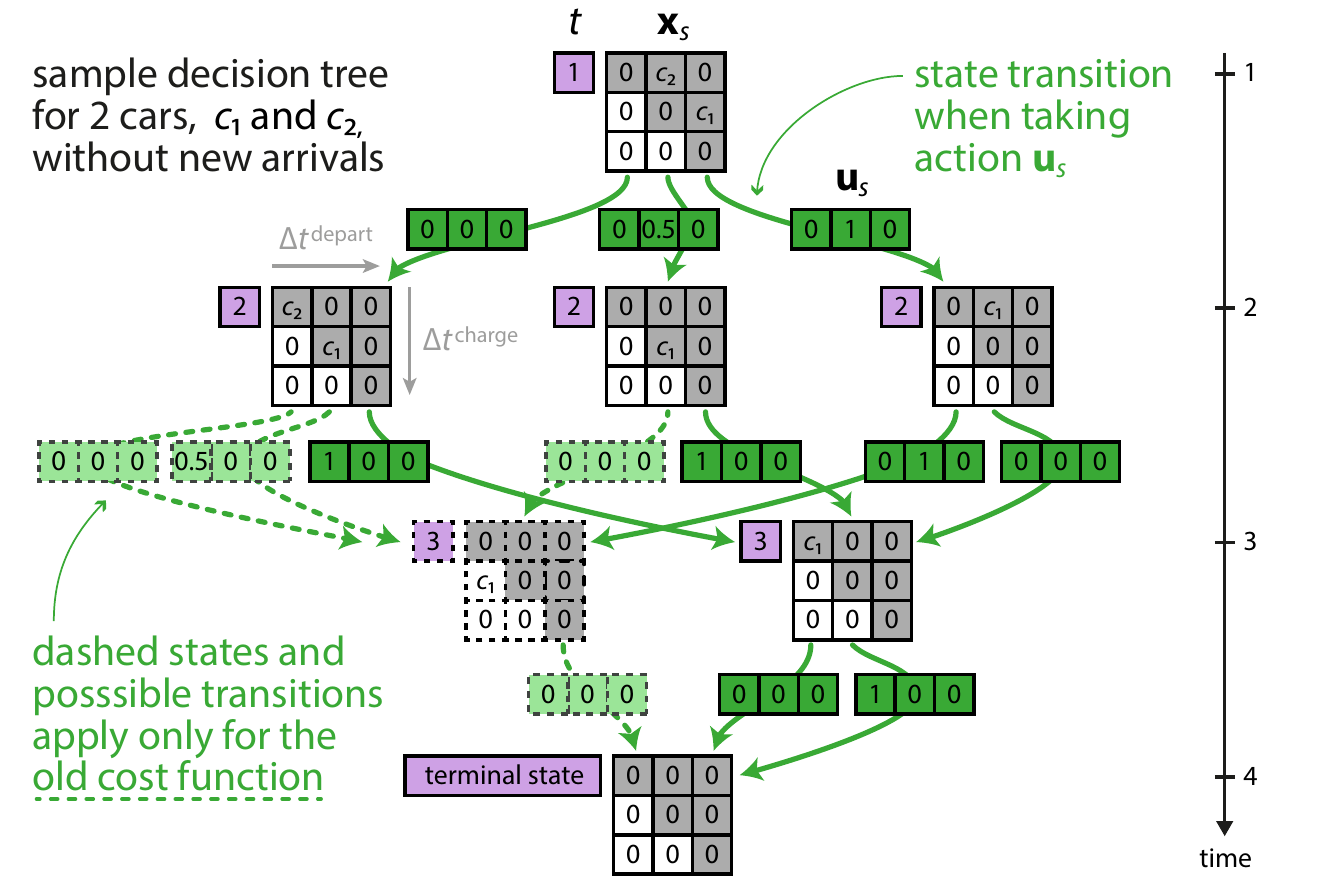}
    	\caption{Example state-action decision tree for $\Nmax = 2$, $\Smax = 3$.}
    	\label{fig:RL:treecompare}
    \end{figure} 
    
    \subsubsection*{Size of State-Action Space:} As opposed to the \textit{action space} for a given state $s$ defined in \cite{nasrin2019journal}, our updated action space is smaller, because we improve the algorithm by not allowing to exploit flexibility where there is none. The total number of possible actions from a given state in the original algorithm (where, for each flexibility $\tflex = d$ we could charge any number of cars between $[0, \textbf{x}^\text{total}_s(d)]$) is:
    
    \begin{equation}\label{eq:RL:oldgrowth}
    \left|\textbf{U}_{s}\right|_\textrm{old} = \prod_{d=0}^{S_{\textit{max}}-1}\left(\textbf{x}_s^\text{total}(d) + 1 \right).
    \end{equation}
    This is updated to:
    \begin{equation}\label{eq:RL:Updatedgrowth}
    \left|\textbf{U}_{s}\right|_\textrm{updated} = 1 + \prod_{d=1}^{S_{\textit{max}}-1}\left(\textbf{x}_s^\text{total}(d) + 1 \right).
    \end{equation}
    The first term in \equref{eq:RL:Updatedgrowth} reflects the single ``choice'' we have for the cars without flexibility (for $d=0$, where $\tflex = 0$). 
    This reduced action space size also shrinks the exploration space for the RL agent (see \secref{subsec:batchRL}). 
    
    \Figref{fig:RL:treecompare} illustrates a scenario of $\Nmax =$ 2 EV charging stations with a horizon of $\Smax =$ 3 slots.
    At time $t =$ 1 we have $N_s$ = 2 connected cars:
    $C_1$ with $(\tdepart_1,\tcharge_1) = (3,2)$, and $C_2$ with $(\tdepart_2,\tcharge_2) = (2,1)$ with no other arrivals during the control horizon. 
    In timeslot 2, the leftmost state in \figref{fig:RL:treecompare} has both cars on the main diagonal, implying 
    they have no flexibility ($\tcharge = \tdepart$, thus $\tflex = 0$) and a fortiori need to be charged.
    Hence, two feasible actions from previous MDP \cite{nasrin2019journal} will not be considered in our updated MDP implementation.

    
    \subsubsection*{Value function:} The learning objective is to minimize the expected $T$-step\footnote{As previously stated, we note the specific control time horizon as $T=\Smax$.} return, which for a policy $\pi$ at timestep $t$ is defined as:
    \begin{equation}\label{eq:RL:Ereturn}
    J^{\pi}_T(s) =\mathbb{E} \left[\sum_{i=t}^{t+T}{C(\underbrace{(t,\textbf{x}_s)}_{s},\textbf{u}_{s},\underbrace{(t+1,\textbf{x}_{s'})}_{s'})}\right]
    \end{equation}
    The policy then amounts to evaluate a state-action value function, and select the action that minimizes it. This value function, commonly named Q-function, is:
    \begin{equation}\label{eq:RL:Qfunc}
    Q^{\pi}(s,\textbf{u}_s) =\mathbb{E} \left[C(s,\textbf{u}_s,s')+J^{\pi}_T(s')\right].
    \end{equation}

\subsection{Batch Reinforcement Learning}
\label{subsec:batchRL}
    We adopt the same batch RL algorithm as in \cite[Algorithm~1]{nasrin2019journal}, fitted Q-iteration (FQI), to approximate $\widehat{Q}^*(s,\textbf{u})$ from past experiences generated using a non-optimum (\eg random) policy.  Each experience is defined in terms of
    \begin{enumerate*}[(i)]
    \item an initial state $s$,
    \item the action taken $\action$, 
    \item the resulting state $s'$ after taking the action, and
    \item the associated costs $C(s,\action,s')$.
    \end{enumerate*}
    The experience set $\mathcal{F}$ is generated based on the cost function used:
    \begin{itemize}
    \item $\mathcal{F}_1$ (\textit{old cost implementation}): Generate all possible actions for a given state, and use the cost function from \equref{eq:RL:OldCost}.
    \item $\mathcal{F}_2$ (\textit{updated cost implementation}): Only allow actions from the updated space-action tree, and use the cost function from \equref{eq:RL:UpdatedCost}.
    \end{itemize}
    Each experience set of $(s,\action,s',C(s,\action,s'))$ tuples is trained separately to deliver an optimum policy $\widehat{Q}^*(s,\textbf{u})$, which is then used evaluated for the testing period. A fully connected Artificial Neural network (ANN) is used as function approximation for $\widehat{Q}^*(s,\textbf{u})$.

\section{EXPERIMENT SETUP}
\label{sec:experimentsetup}

Our simulations will compare the original (see \cite{nasrin2019journal}) cost function and associated state-action space, with the updated ones, in terms of the resulting  policies, trained with the respective updated and old cost datasets ($\mathcal{F}_1$ and $\mathcal{F}_2$).
Experiments are run on a system with an Intel Xeon E5645 3.1\,GHz processor and 16\,GB RAM. 

\subsection{Parameters and Settings}

\label{subsec:algosettings}
    \subsubsection*{Data preparation:}
    Our dataset is derived from real world data collected by ElaadNL since 2011, from 2500+ public charging stations \cite{nasrin2018data}.
    The maximum connection duration is set to $\Hmax = 24$\,h with time granularity  $\tslot = 2$\,h, which means $\Smax$ = 12 timeslots. 
    We jointly coordinate $\Nmax =10$ charging stations. 

    As summarized in \secref{subsec:batchRL}, the input to the policy learning is a set of experiences, which depends on the cost function used: $\mathcal{F}_1$ (\textit{old cost implementation}) and $\mathcal{F}_2$ (\textit{updated cost implementation}).
    These experiences are collected as so-called trajectories by beginning at the starting of the day ($t_1$,$x_1$), taking actions randomly until a terminal state is reached (for details, see \cite{nasrin2019journal}). Each such trajectory 
    comprises 12 tuples $(s_i,\textbf{u}_{s,i},s_{i+1},C(s_i,\textbf{u}_{s,i},s_{i+1}))$. We vary the number of unique trajectories per day, $N_\text{traj} \in \{ 5K, 10K, 15K, 20K \}$.
    
   
\subsubsection*{Function Approximator:} We use the same artificial neural network (ANN) architecture as in \cite{nasrin2019journal}, comprising an input layer and 2 hidden layers with ReLU activation functions.

\subsection{Evaluation}
\label{subsec:evaluations}
    
    \subsubsection*{Training time:} Defined as the time it takes for RL agent to be trained. To generate training data, we take contiguous period for a given duration of $\Delta t \in \{1, 3, 5, 7, 9\ \text{months}\}$. For each $\Delta t$, we randomly selected 5 training data periods, each within the range between Jan.\ 1, 2015 and Sep.\ 30, 2015. We run 12 iterations to train the ANN for each selected period for a given $\Delta t$. We record the training time for each of these training datasets, for both agents (one trained on $\mathcal{F}_1$ and the other on $\mathcal{F}_2$). 
    
    \subsubsection*{Cost comparison:} The last 3 months of 2015 are used as the test set for evaluation, \ie $\mathcal{B}^{\text{test}}=\{e_i|i = 274,\dots,365\}$  containing 92 days. 
    We report the same normalized cost as in \cite{nasrin2019journal}, given by:
    \begin{equation}\label{eq:RL:measure}
    C_{\pi(\Delta t, j)}=\frac{1}{|\mathcal{B}^{\text{test}}|}\sum_{e\in \mathcal{B}^{\text{test}}}\frac{C_{\pi(\Delta t, j)}^e}{C^e_{\text{opt}}}.
    \end{equation} 
    Following costs are calculated and compared to 
    analyze the various policies:
    \begin{itemize}
    \item $\CRLnew$: cost of the policy trained with the updated cost.
    \item $\CRLold$: cost of the policy obtained by \cite{nasrin2019journal}.
    \item $\Cbau$: cost of the business-as-usual (BAU) policy\footnote{Continuously charge each EV upon arrival.}.
    \item $\Copt$: for an optimum policy, derived from optimization with perfect knowledge of future EV connections.
    \item $\Cheur$: for a discrete-action heuristic.
    \end{itemize}
    The latter heuristic policy assumes that individual EVs are charged uniformally over their entire connection time.\footnote{Specifically, the heuristic spreads the $c$ slots that the EV needs to charge over the total available number of slots $d$. This amounts to distributing $d - c$ no-charge slots evenly over the total number of $d$ slots, thus splitting them into $d - c + 1$ parts. Assuming for simplicity that $c \geq d/2$, this means we insert a no-charge slot every $\floor*{d / (d - c + 1)}$ other slots. (For $c < d/2$, similarly distribute `charge' slots evenly over the majority of `no-charge' slots.)}

\section{Results}
\label{sec:results}
\subsection{Training Time}
\label{subsec:trainingtime}

\begin{figure}[!t]
        \centering
        \includegraphics[width=\columnwidth]{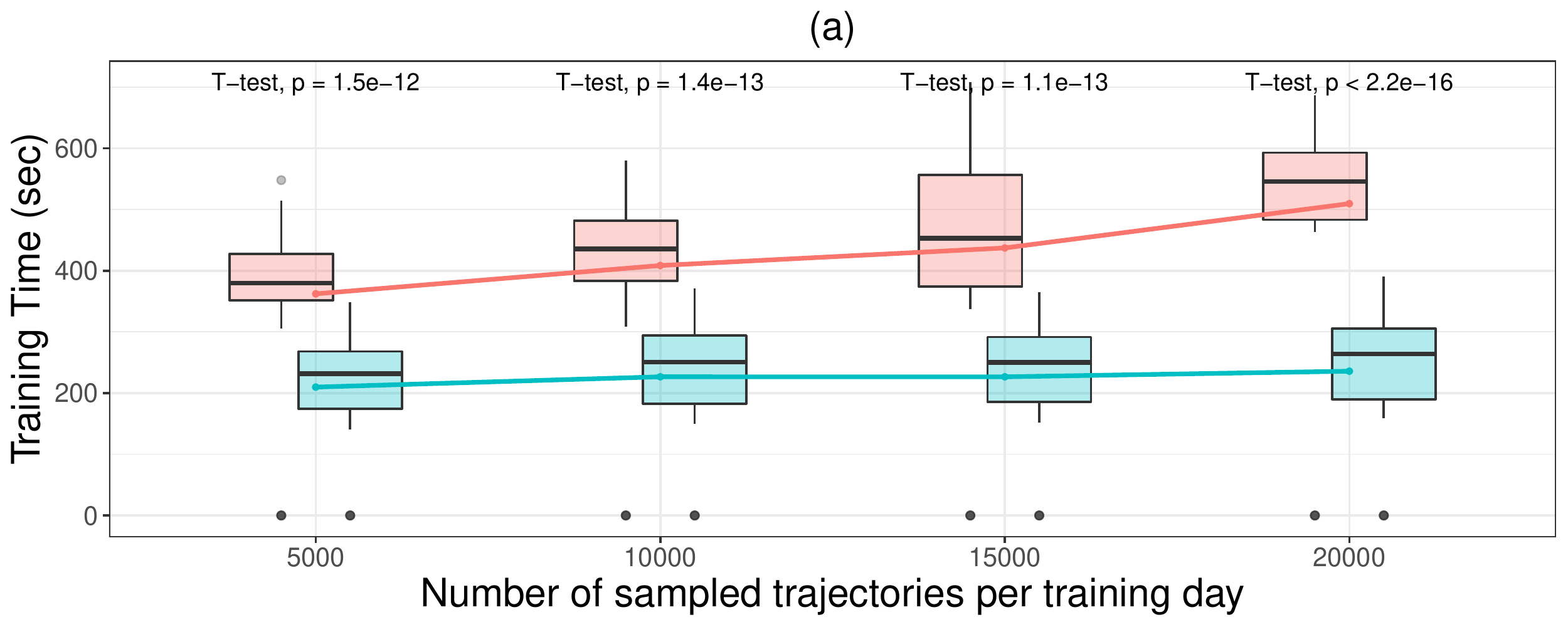}
        \includegraphics[width=\columnwidth]{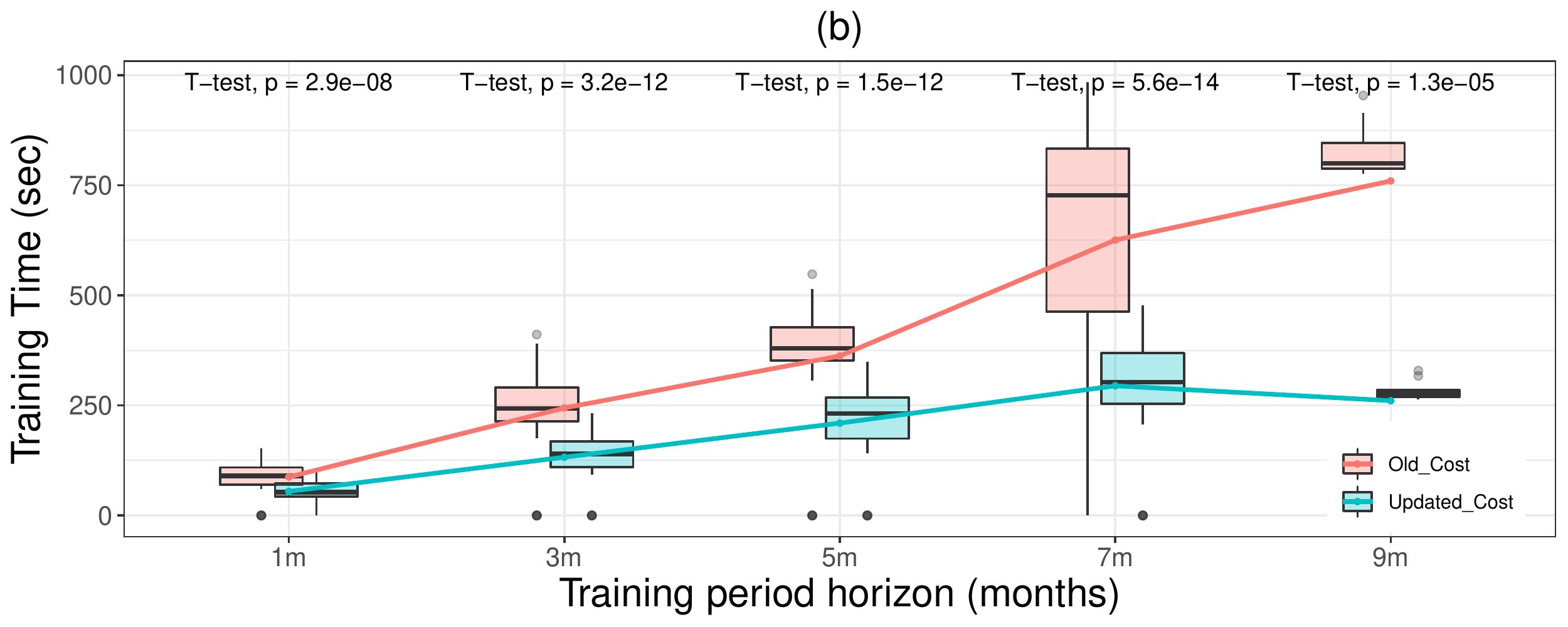}
        \caption{Policy training time for varying (a)~number of samples per day $N_\text{traj}$ (for $\Delta t$ = 5 months),  and (b)~training period horizon $\Delta t$ (for $N_\text{traj}$ = 5000) .}
        \label{fig:trainingtime:comp}
    \end{figure}

    \Figref{fig:trainingtime:comp} 
    shows the training time for each of the old and new policies, \ie using the respective cost functions \equref{eq:RL:OldCost} and \equref{eq:RL:UpdatedCost}, to answer research questions \textbf{\qref{q:traintime:reduction}} (\ie what training time reduce does the update cost function achieve?) and \textbf{\qref{q:triantime:variance}} (\ie how do training set parameters affect training time?). \Figref{fig:trainingtime:comp}(a) compares the training time for increasing training dataset size in terms of number of sampled trajectories, 
    for a training period of $\Delta t = 5$ months.\footnote{We noted a similar trend for all time spans, \ie for all $\Delta t \in \{1, 3, 5, 7, 9 \text{ months}\}$.} We note that our updated cost function and resulting policy achieves a reduction of training time compared to the old ones (from~\cite{nasrin2019journal}) in the range of 42\%--54\% (for 5k--20k sampled trajectories per training day; averages over 5 runs).
    
    \Figref{fig:trainingtime:comp}(b) reports similar relative differences in training time between old and updated policies when varying training period time spans, \ie $\Delta t \in \{1, 3, 5, 7, 9 \text{ months}\}$. Training time savings now range from 37\%--53\% when varying $\Delta t$ from 1 to 7 months.

    

\subsection{Cost Reduction Comparison}
\label{subsec:performance}
    \begin{figure}[!t]
        \centering
        \includegraphics[width=\columnwidth]{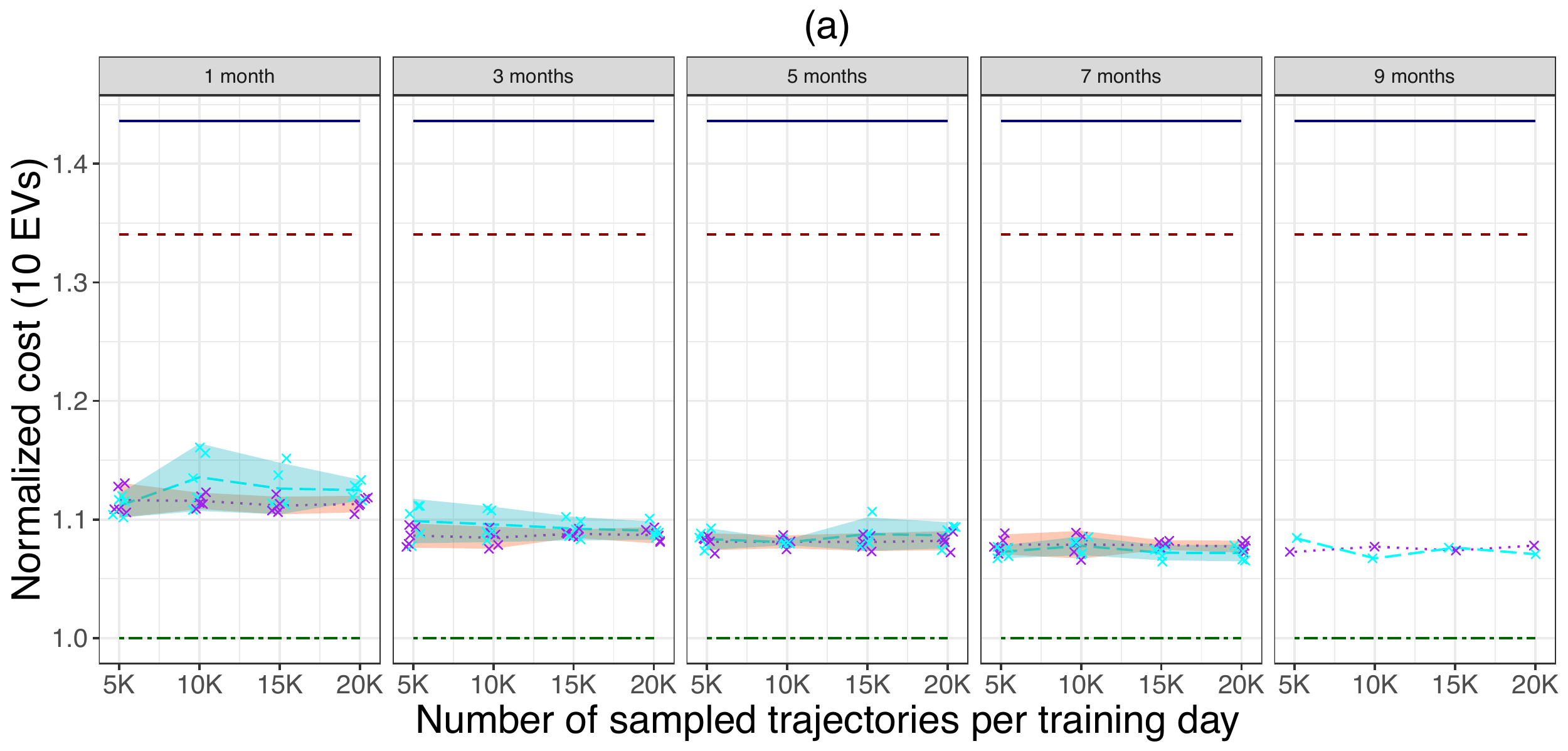}
        \includegraphics[width=\columnwidth]{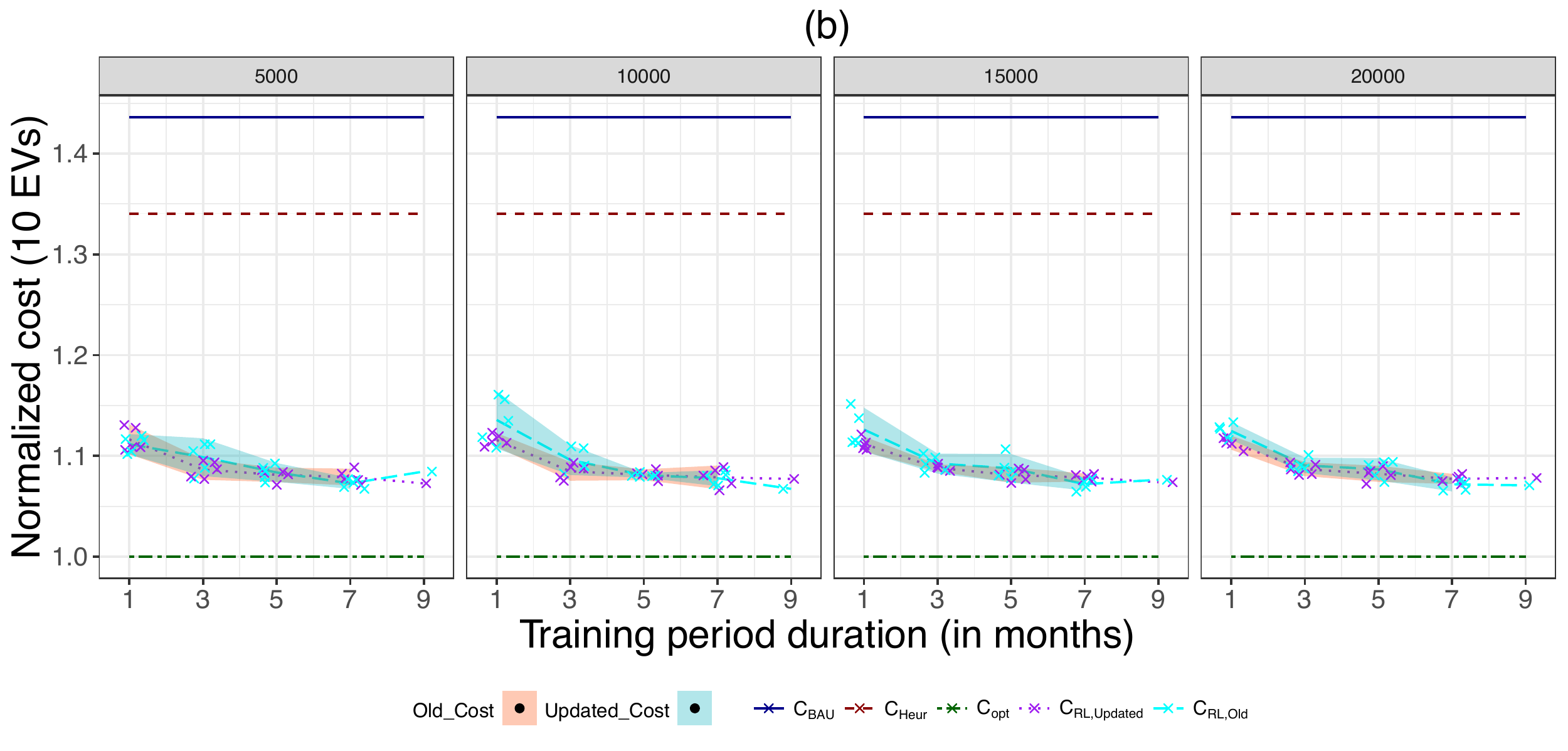}
        \caption{Normalized costs for increasing
        (a)~$N_\text{traj}$ and (b)~$\Delta t$.}
        \label{fig:cost:comp}
    \end{figure}
    
    \Figref{fig:cost:comp} compares the normalized cost achieved by our improved formulation with that of previous work~\cite{nasrin2019journal} as well as the baselines from \secref{subsec:evaluations}.
    We note how normalized costs for the various RL policies vary with increasing training sets, in terms of 
    \begin{enumerate*}[(a)] 
    \item number of sampled trajectories per training day, and
    \item training period time spans.
    \end{enumerate*}
    Comparing old and new cost policies, we note no significant difference in the normalized cost (\ie $C_\text{RL,updated} \approx C_\text{RL,old}$). There are slight variations in the distributions, which are statistically insignificant, and may be caused by the randomization of the samples. It is also worth noting that the RL algorithm performs better than both business-as-usual (BAU) and the newly defined heuristic policy. Unexpectedly, it performs worse the the all-knowing optimum policy baseline, since the latter has exact visibility of future EV arrivals. Still, the RL approach deviates less than 10\%. 
    
    We conclude for \textbf{\qref{q:performance}} (effect of updating the RL cost function on the RL policy?) that performance of both cost functions is statistically similar (over multiple training sets).

\subsection{Flexibility Utilization}
\label{subsec:ETflex}
    \begin{figure}[!t]
        \centering
        \includegraphics[width=\columnwidth]{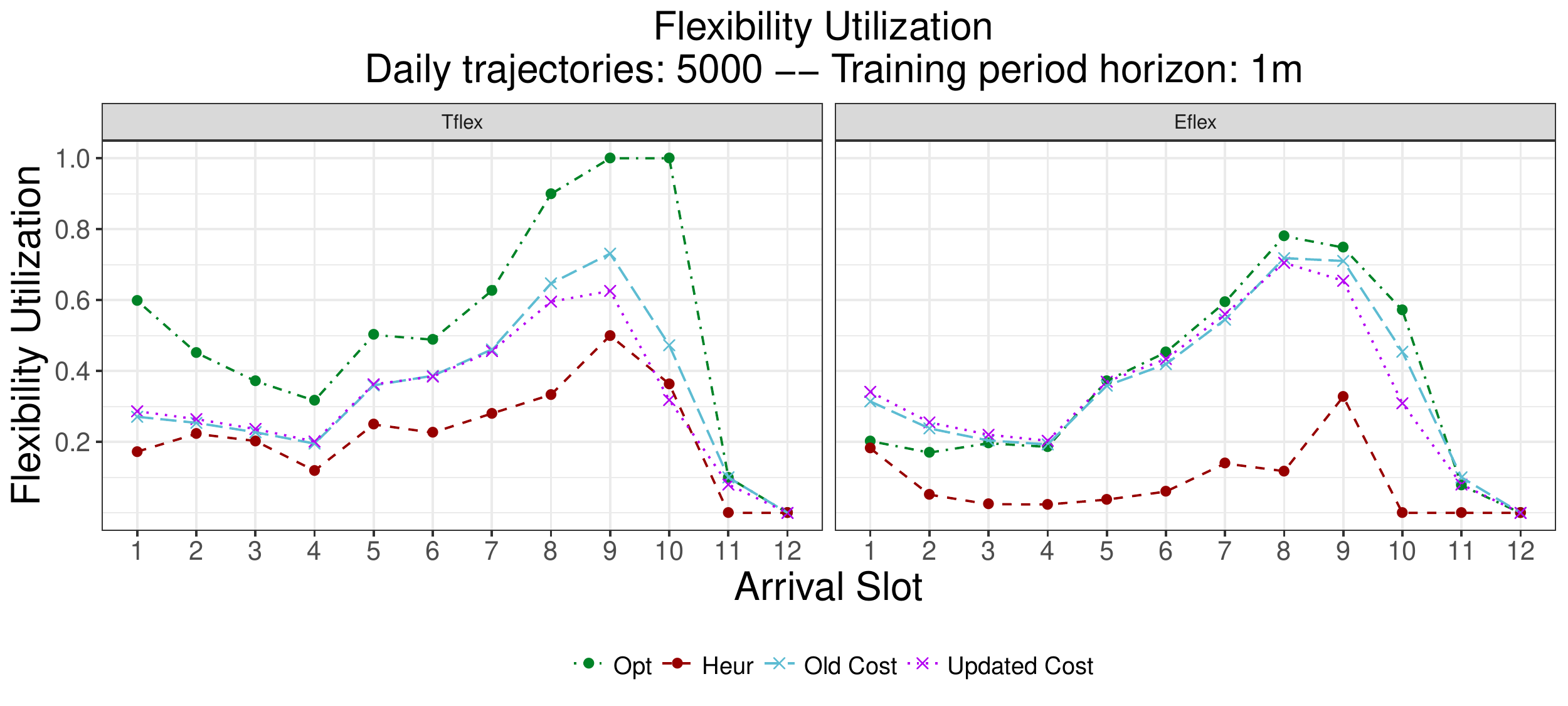}
        \caption{Flexibility Utilization for cars arrived in each slot.}
        \label{fig:et:flex}
    \end{figure} 
    To quantify the overall utilization of flexibility offered by users, we use previously introduced energy ($\Eflex$) and time flexibility ($\Tflex$) measures in \cite{nasrin2018data}.
    These measures are important for energy providers and users alike (\eg for providing incentives, cost minimization, time management). 
    In short, $\Eflex$ reports the fraction of total charging load that is shiftable outside the BAU charging interval that is effectively deferred, and $\Tflex$ the fraction of the flex time window ($\tflex$) that is actually exploited.
    
    To answer \textbf{\qref{q:etflex}}, \figref{fig:et:flex} plots $\Tflex$ and $\Eflex$ for EVs for each timeslot they arrive in, thus quantifying the utilized flexibility for the exemplary\footnote{We notice similar trends in all sample sizes and time spans.} training data of $N_\text{traj} =$ 5K sample trajectories per day, and period $\Delta t =$ 1 month. We note that the RL learning policy utilizes much more flexibility than the heuristic policy. On average, 40\% of provided energy flexibility is utilized with RL. It is also important to note that RL policies energy flexibility utilization is close to that exploited in the all-knowing optimum policy. This indicates the trained RL agent balances loads similarly as the optimum policy, despite having no a priori exact knowledge of future EV arrivals.

\section{Conclusion}
\label{sec:conclusion}
    
    
    We significantly improve the previously proposed reinforcement learning (RL) strategy in \cite{nasrin2019journal}, to learn a policy coordinating the charging of multiple EVs. Our updated MDP definition, with a new cost function, strongly shrinks the the state-action space and thus significantly reduces training time (with $> 50\%)$, while retaining the beneficial performance of the RL trained policy. This training time reduction increases with larger training sets, thus suggesting the updated cost approach to be more practical.
    Future work includes evaluating an $\epsilon$-greedy RL approach for the new MDP definition.

\begin{acks}
The authors thank Joachim van der Herten for his feedback on \cite{nasrin2019journal} that inspired the current work. 
\end{acks}

\bibliographystyle{ACM-Reference-Format}
\ifanon
\bibliography{references,references-own-anonymous}
\else
\bibliography{references,references-own}


\begin{thebibliography}{00}


\ifx \showCODEN    \undefined \def \showCODEN     #1{\unskip}     \fi
\ifx \showDOI      \undefined \def \showDOI       #1{#1}\fi
\ifx \showISBNx    \undefined \def \showISBNx     #1{\unskip}     \fi
\ifx \showISBNxiii \undefined \def \showISBNxiii  #1{\unskip}     \fi
\ifx \showISSN     \undefined \def \showISSN      #1{\unskip}     \fi
\ifx \showLCCN     \undefined \def \showLCCN      #1{\unskip}     \fi
\ifx \shownote     \undefined \def \shownote      #1{#1}          \fi
\ifx \showarticletitle \undefined \def \showarticletitle #1{#1}   \fi
\ifx \showURL      \undefined \def \showURL       {\relax}        \fi
\providecommand\bibfield[2]{#2}
\providecommand\bibinfo[2]{#2}
\providecommand\natexlab[1]{#1}
\providecommand\showeprint[2][]{arXiv:#2}

\bibitem[\protect\citeauthoryear{Chi{\c{s}}, Lund{\'e}n, and
  Koivunen}{Chi{\c{s}} et~al\mbox{.}}{2016}]%
        {chics2016reinforcement}
\bibfield{author}{\bibinfo{person}{Adriana Chi{\c{s}}}, \bibinfo{person}{Jarmo
  Lund{\'e}n}, {and} \bibinfo{person}{Visa Koivunen}.}
  \bibinfo{year}{2016}\natexlab{}.
\newblock \showarticletitle{Reinforcement learning-based plug-in electric
  vehicle charging with forecasted price}.
\newblock \bibinfo{journal}{{\em IEEE Trans. Vehic. Technol.\/}}
  \bibinfo{volume}{66}, \bibinfo{number}{5} (\bibinfo{year}{2016}),
  \bibinfo{pages}{3674--3684}.
\newblock


\bibitem[\protect\citeauthoryear{Claessens, Vandael, Ruelens, De~Craemer, and
  Beusen}{Claessens et~al\mbox{.}}{2013}]%
        {claessens2013RLEV}
\bibfield{author}{\bibinfo{person}{Bert~J. Claessens}, \bibinfo{person}{Stijn
  Vandael}, \bibinfo{person}{Frederik Ruelens}, \bibinfo{person}{Klaas
  De~Craemer}, {and} \bibinfo{person}{Bart Beusen}.}
  \bibinfo{year}{2013}\natexlab{}.
\newblock \showarticletitle{Peak shaving of a heterogeneous cluster of
  residential flexibility carriers using reinforcement learning}. In
  \bibinfo{booktitle}{{\em IEEE PES ISGT Europe 2013}}.
  \bibinfo{publisher}{IEEE}, \bibinfo{pages}{1--5}.
\newblock


\bibitem[\protect\citeauthoryear{Hu, Morais, Sousa, and Lind}{Hu
  et~al\mbox{.}}{2016}]%
        {hu2016electric}
\bibfield{author}{\bibinfo{person}{Junjie Hu}, \bibinfo{person}{Hugo Morais},
  \bibinfo{person}{Tiago Sousa}, {and} \bibinfo{person}{Morten Lind}.}
  \bibinfo{year}{2016}\natexlab{}.
\newblock \showarticletitle{Electric vehicle fleet management in smart grids: A
  review of services, optimization and control aspects}.
\newblock \bibinfo{journal}{{\em Renew. Sust. Energ. Rev.\/}}
  \bibinfo{volume}{56} (\bibinfo{year}{2016}), \bibinfo{pages}{1207--1226}.
\newblock


\bibitem[\protect\citeauthoryear{Sadeghianpourhamami, Deleu, and
  Develder}{Sadeghianpourhamami et~al\mbox{.}}{2018a}]%
        {nasrin2018mdp}
\bibfield{author}{\bibinfo{person}{Nasrin Sadeghianpourhamami},
  \bibinfo{person}{Johannes Deleu}, {and} \bibinfo{person}{Chris Develder}.}
  \bibinfo{year}{2018}\natexlab{a}.
\newblock \showarticletitle{Achieving Scalable Model-Free Demand Response in
  Charging an Electric Vehicle Fleet with Reinforcement Learning}. In
  \bibinfo{booktitle}{{\em e-Energy2018, the 9th ACM International Conference
  on Future Energy Systems}}. ACM Press, \bibinfo{pages}{1--3}.
\newblock


\bibitem[\protect\citeauthoryear{Sadeghianpourhamami, Deleu, and
  Develder}{Sadeghianpourhamami et~al\mbox{.}}{2019}]%
        {nasrin2019journal}
\bibfield{author}{\bibinfo{person}{Nasrin Sadeghianpourhamami},
  \bibinfo{person}{Johannes Deleu}, {and} \bibinfo{person}{Chris Develder}.}
  \bibinfo{year}{2019}\natexlab{}.
\newblock \showarticletitle{Definition and evaluation of model-free
  coordination of electrical vehicle charging with reinforcement learning}.
\newblock \bibinfo{journal}{{\em IEEE Transactions on Smart Grid\/}}
  (\bibinfo{year}{2019}).
\newblock


\bibitem[\protect\citeauthoryear{Sadeghianpourhamami, Refa, Strobbe, and
  Develder}{Sadeghianpourhamami et~al\mbox{.}}{2018b}]%
        {nasrin2018data}
\bibfield{author}{\bibinfo{person}{Nasrin Sadeghianpourhamami},
  \bibinfo{person}{Nazir Refa}, \bibinfo{person}{Matthias Strobbe}, {and}
  \bibinfo{person}{Chris Develder}.} \bibinfo{year}{2018}\natexlab{b}.
\newblock \showarticletitle{Quantitive analysis of electric vehicle
  flexibility: A data-driven approach}.
\newblock \bibinfo{journal}{{\em International Journal of Electrical Power \&
  Energy Systems\/}}  \bibinfo{volume}{95} (\bibinfo{year}{2018}),
  \bibinfo{pages}{451--462}.
\newblock


\bibitem[\protect\citeauthoryear{V\'{a}zquez-Canteli and
  Nagy}{V\'{a}zquez-Canteli and Nagy}{2019}]%
        {VazquezCanteli2019}
\bibfield{author}{\bibinfo{person}{Jos\'{e}~R. V\'{a}zquez-Canteli} {and}
  \bibinfo{person}{Zolt\'{a}n Nagy}.} \bibinfo{year}{2019}\natexlab{}.
\newblock \showarticletitle{Reinforcement learning for demand response: {A}
  review of algorithms and modeling techniques}.
\newblock \bibinfo{journal}{{\em Appl. Energ.\/}}  \bibinfo{volume}{235}
  (\bibinfo{year}{2019}), \bibinfo{pages}{1072--1089}.
\newblock
\showISSN{0306-2619}


\end{thebibliography}
\fi



\end{document}